\documentclass[10pt,twocolumn,letterpaper]{article}

\usepackage[pagenumbers]{iccv} 

\usepackage{algorithm}
\usepackage{algorithmic}
\usepackage{multirow}
\usepackage{graphicx}

\definecolor{iccvblue}{rgb}{0.21,0.49,0.74}
\usepackage[pagebackref,breaklinks,colorlinks,allcolors=iccvblue]{hyperref}

\def\paperID{7206} 
\def\confName{ICCV}
\def\confYear{2025}

\title{Classifying Long-tailed and Label-noise Data via Disentangling and Unlearning}

\author{
Shu Chen\textsuperscript{\rm 1} \quad Mengke Li\textsuperscript{\rm 2} \quad Yiqun Zhang\textsuperscript{\rm 3} \quad Yang Lu\textsuperscript{\rm 1} \quad Bo Han\textsuperscript{\rm 4} \quad
Yiu-ming Cheung\textsuperscript{\rm 4} \quad Hanzi Wang\textsuperscript{\rm 1} \\
{\textsuperscript{\rm 1}Xiamen University} \quad
{\textsuperscript{\rm 2}Shenzhen University} \quad
{\textsuperscript{\rm 3}Guangdong University of Technology} \\
{\textsuperscript{\rm 4}Hong Kong Baptist University}\\
}

\begin{document}
\maketitle

\begin{abstract}
In real-world datasets, the challenges of long-tailed distributions and noisy labels often coexist, posing obstacles to the model training and performance. Existing studies on long-tailed noisy label learning (LTNLL) typically assume that the generation of noisy labels is independent of the long-tailed distribution, which may not be true from a practical perspective. 
In real-world situaiton, we observe that the tail class samples are more likely to be mislabeled as head, exacerbating the original degree of imbalance. We call this phenomenon as ``tail-to-head (T2H)'' noise.
T2H noise severely degrades model performance by polluting the head classes and forcing the model to learn the tail samples as head. 
To address this challenge, we investigate the dynamic misleading process of the nosiy labels and propose a novel method called Disentangling and Unlearning for Long-tailed and Label-noisy data (DULL). It first employs the Inner-Feature Disentangling (IFD) to disentangle feature internally.  Based on this, the Inner-Feature Partial Unlearning (IFPU) is then applied to weaken and unlearn incorrect feature regions correlated to wrong classes. This method prevents the model from being misled by noisy labels, enhancing the model's robustness against noise.
To provide a controlled experimental environment, we further propose a new noise addition algorithm to simulate T2H noise.
Extensive experiments on both simulated and real-world datasets demonstrate the effectiveness of our proposed method. Our code is available at \url{https://anonymous.4open.science/r/DULL-E222}.
\end{abstract}

\section{Introduction}
\label{sec:intro}
The long-tail problem is a significant challenge in machine learning, focusing on mitigating the decline in model performance on tail classes \cite{nitesh2002smote, zhang2021distribution, li2022long, kim2020m2m, kang2019decoupling}. In real-world datasets, long-tailed distributions often coexist with noisy labels. Recently, the long-tailed noisy label learning (LTNLL) has gained increasing attention. Existing LTNLL research mainly assumes that the noise ratios are identical across all classes and focuses on how to separate clean and noisy labels in tail \cite{wei2021robust, huang2022uncertainty, cao2020heteroskedastic, zhang2023noisy}. 

\begin{figure}[t]
  \centering
    \begin{subfigure}{1\linewidth}
        \includegraphics[width=\linewidth]{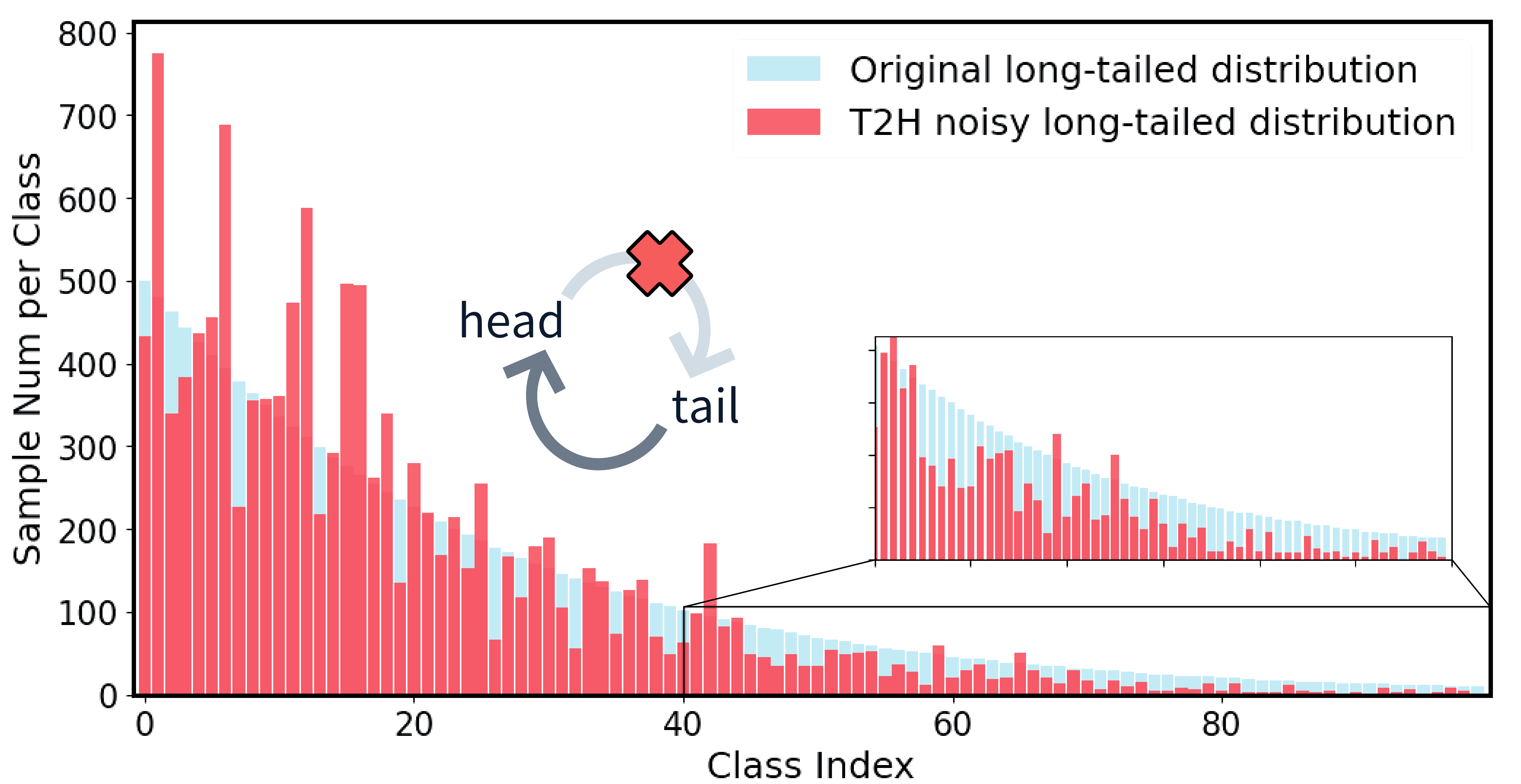}
        \caption{The original and noisy long-tailed distributions}
        \label{fig:description1}
    \end{subfigure}
    \hfill
    \begin{subfigure}{0.43\linewidth}
        \includegraphics[width=0.9\linewidth]{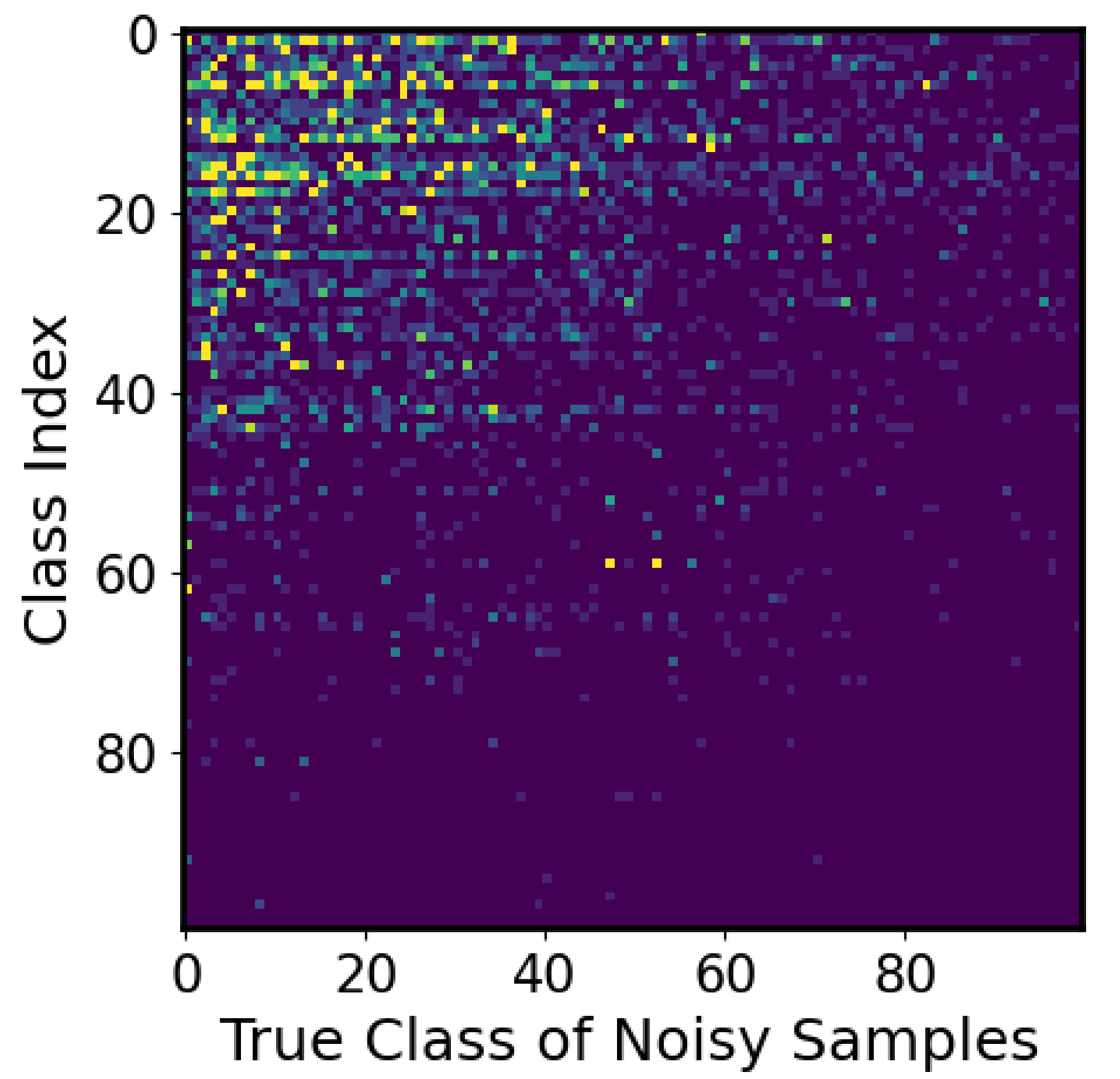}
        \caption{Noisy label composition per class}
        \label{fig:description2}
    \end{subfigure}
    \hspace{2pt}
    \begin{subfigure}{0.50\linewidth}
        \includegraphics[width=\linewidth]{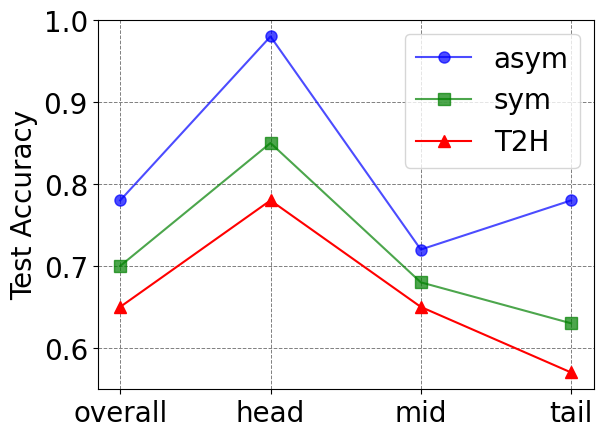}
        \caption{The impact of different types of noise labels}
        \vspace{3pt}
        \label{fig:damage1}
    \end{subfigure}

   \caption{A case study of tail-to-head (T2H) noisy and long-tailed distribution of CIFAR-100 with an original imbalance factor of 10. 
   }
   \label{fig:onecol}
\end{figure}

\begin{figure*}[!t] 
  \centering
  \begin{subfigure}{0.67\linewidth} 
    \includegraphics[width=\linewidth]{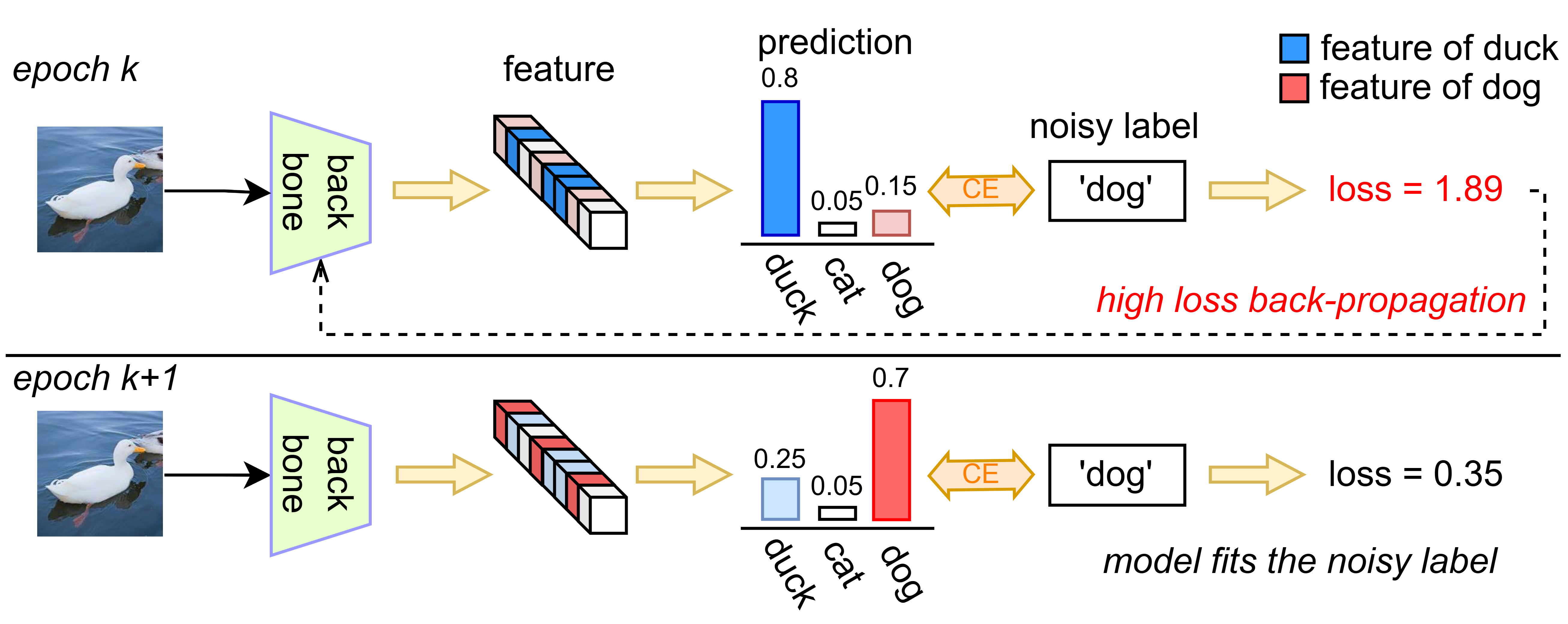} 
    \caption{The dynamic process in which noisy labels cause the model to reinforce feature representations of incorrect classes.} 
    \label{fig:motivation1}
  \end{subfigure}
  \hspace{0.5cm} 
  \begin{subfigure}{0.22\linewidth}
    \includegraphics[width=\linewidth]{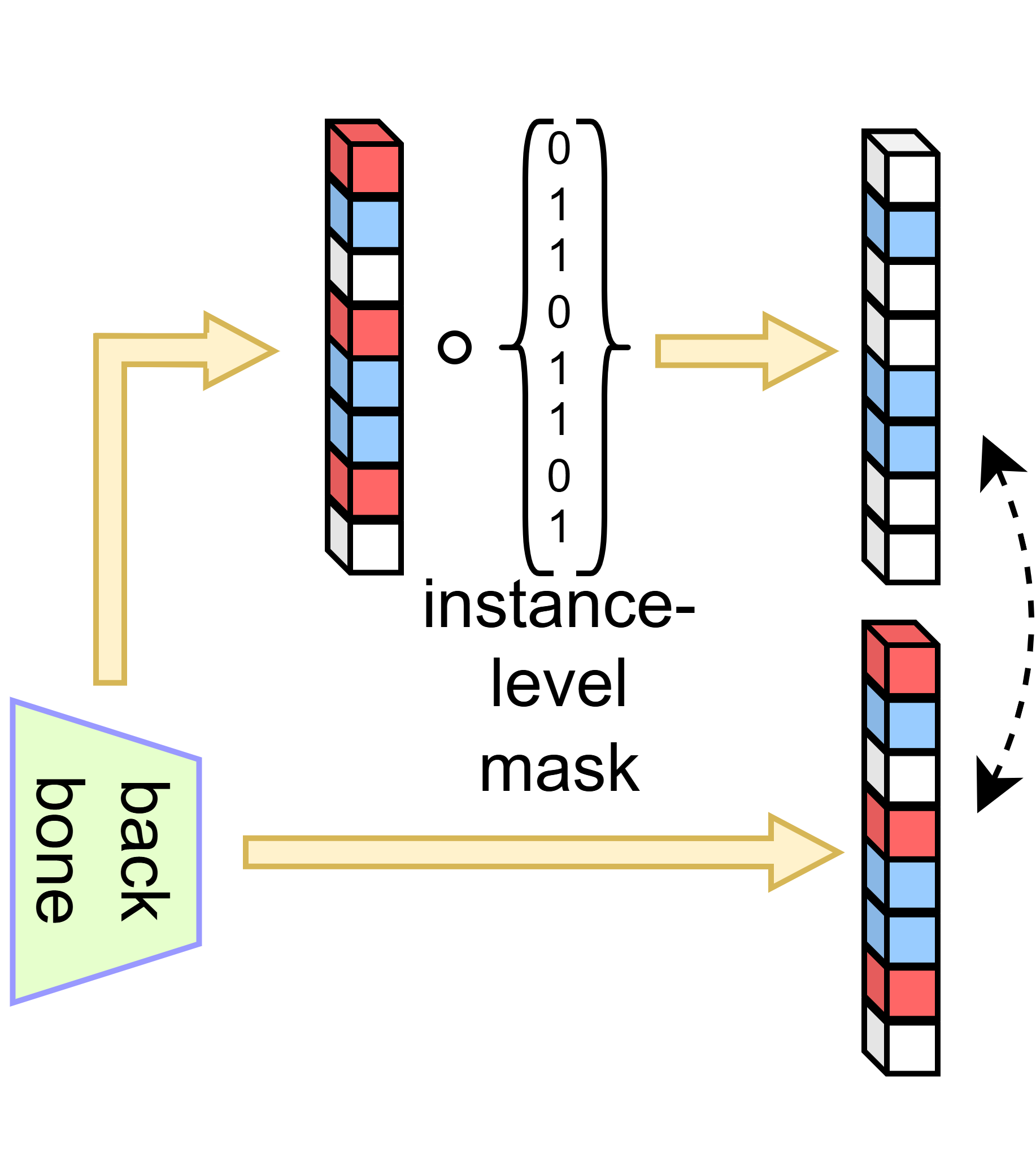}
    \caption{The core of the IFPU mechanism illustration} 
    \label{fig:motivation2}
  \end{subfigure}
  \caption{(a) In epoch \(k\), the model extracts salient duck features (blue regions), producing higher prediction in duck class. However, the prediction and the noisy label `dog' result in a high loss, which adjusts the model to reinforce and output salient dog feature by back-propagation. In epoch \(k+1\), the updated model outputs more salient dog features (red regions) and less salient duck features for the same duck sample. This illustrates how the model is misled by noisy labels, leading to a degradation in classification performance. (b) Illustration of the core of IFPU mechanism, show how it selectively unlearns incorrect feature regions associations to wrong classes, preventing model's wrong reinforcement, thereby enhancing robustness against noisy data.}
  \label{fig:cross-column}
\end{figure*}

There is an implicit assumption that the generation of noisy labels is independent of the long-tailed distribution. However, in real-world situations, we observe that the long-tail problem and noisy labels problem are non-orthogonal and interact with each other. Specifically, tail samples are more likely to be mislabeled as head samples by annotators due to the scarcity, while head samples are less likely to be mislabeled. This unidirectional mislabeling tendency further exacerbates the imbalance of the long-tailed distribution. In summary, the long-tailed distribution promotes the generation of noisy labels, while noisy labels in turn exacerbate the long-tailed imbalance, creating a mutually deteriorating relationship. We call this phenomenon ``tail-to-head (T2H)'' noise. A real-world case is shown in Fig.~\ref{fig:description1}. T2H noise widely exists in real-world situations. For example, in long-tailed medical data, like ChestX-ray14 \cite{wang2017chestxray8}, rare diseases are often misdiagnosed as common ones due to their infrequency (e.g., pneumothorax misdiagnosed as pneumonia). Compared to traditional long-tailed noise, T2H noise has the following unique characteristics:
(1) Unidirectional tendency for noise generation from tail classes to head classes.
(2) T2H changes and exacerbates the original long-tailed distribution; 
(3) T2H leads to varying noise ratios across classes, as shown in Fig.~\ref{fig:description2};
(4) T2H severely degrades model performance, as shown in Fig.~\ref{fig:damage1}.

The cause of the performance degradation by T2H noise is mainly in the pollution of head-class, as well as the knowledge misguidance suffered by tail classes. Specifically, the presence of numerous noise samples in head leads to the pollution of the head-class feature space and supervision labels. This makes it difficult for the model to capture the core features of head classes, thereby degrading the classification performance on head. Meanwhile, tail classes, which are already scarce in samples, face an greater learning challenge due to the reduction in effective sample size caused by T2H noise. More critically, tail-class noise samples that are mislabeled as head force the model to associate tail-class features with head. This misassociation misguides the model to incorrectly classify tail-class samples as head. We investigate the dynamic process of how models are misled by noisy labels during learning. As shown in Fig.~\ref{fig:motivation1}, when the model learns a sample that is actually "duck" but is mislabeled as ``dog," the noisy label forces the model to output features similar to ``dog" by back-propagation of high loss. This learning process of noisy labels reinforces the feature representation of the incorrect class, thereby generating the misguidance.


\begin{figure*}[htbp]
	\center{\includegraphics[width=\textwidth]{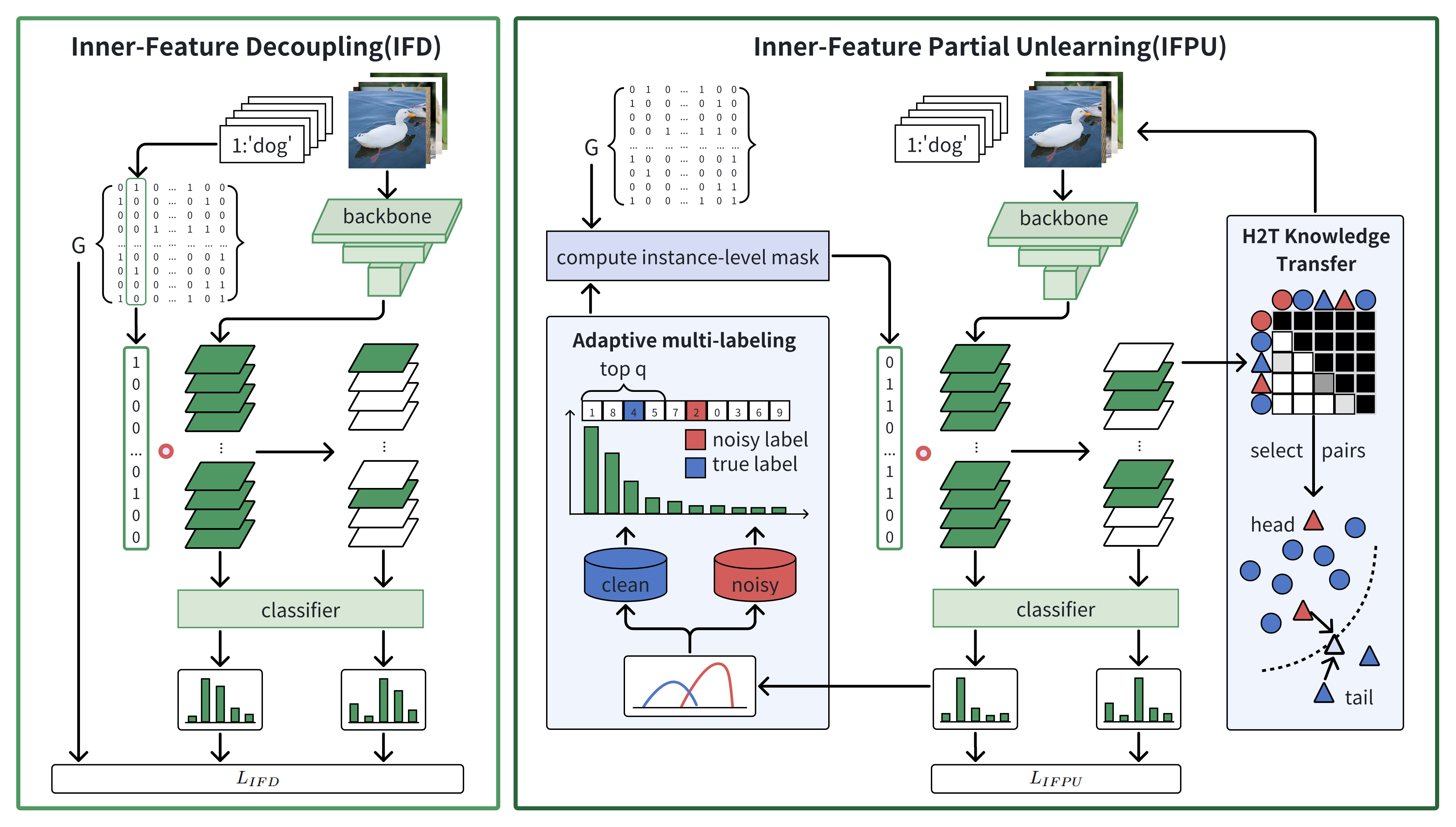}} 
	\caption{Illustration of our proposed DULL method. The left panel shows the Inner-Feature Disentangling (IFD) mechanism, which aims to separate feature channels into independent regions, ensuring that deactivating one feature region does not impact others. Based on the disentangled features, the Inner-Feature Partial Unlearning (IFPU) mechanism, illustrated in the right panel, unlearns incorrect feature regions associated with wrong classes, thereby preventing the model from reinforcing incorrect information.}
	\label{method_all}
\end{figure*}


To tackle the issues, we propose a novel method, called Disentangling and Unlearning for Long-tailed and Label-noisy data (DULL) to weaken the incorrect feature reinforcement. 
It contains two core mechanisms: Inner-Feature Disentangling (IFD) and Inner-Feature Partial Unlearning (IFPU). 
IFD aims to disentangle the class-related channels within individual feature. By orthogonalization, IFD disentangle each feature channels into independent regions, ensuring that each channel associated with only one class. This process lays the foundation for IFPU, allowing the selective unlearning of incorrect features regions without disrupting the learning of other regions.
After disentangling features internally, IFPU performs selective unlearning on each sample's features. Specifically, IFPU identifies and zeroes out the feature regions associated with wrong classes. This weakens the model's reinforcement of these incorrect features regions and achieves ``unlearning'' of them, alleviating the knowledge misguidance caused by noisy labels.
The main contributions of our research are as follows:
\begin{itemize}
    \item We observe and study a novel and challenging noisy labels problem combined with long-tailed data, called T2H noise.
    
    \item We investigate the process in which noisy labels mislead models into reinforcing incorrect features regions.
    
    \item We introduce a novel method, DULL, which integrates disentangling and unlearning.  Extensive experiments have demonstrated the effectiveness of our method.
\end{itemize}

\section{Problem setup}

We assume a long-tailed dataset \(\tilde{\mathcal{D}} = \{(x_i, \tilde{y}_i)\}_{i=1}^{N}\) with \(C\) class, where \(x_i\) is the $i$-th instance, \(\tilde{y}_i \in C\) is the true label of \(x_i\), and \(N\) is the total number of instances.
For each class \(k\), \(\tilde{\mathcal{S}}_k \in \tilde{\mathcal{D}}\) denotes the set of instances truly belonging to class \(k\). \(\tilde{N}_k := |\tilde{\mathcal{S}}_k|\) represent the number of the instances of class \(k\), which have \(\tilde{N}_1 \geq \tilde{N}_2 \geq \cdots \geq \tilde{N}_c \). For each instance \(x_i\), \(y_i \in C\) denotes its observed label, which may be incorrect due to label noise. The presence of label noise changes the observed counts of instances across classes, resulting in a shifted distribution of the observed data. In this paper, we consider T2H noise that causes a unidirectional transfer of tail class instances to head classes, resulting in an observed increase of head class instances and a shortage of tail class instances, further exacerbating the imbalance. Define a transition matrix \(T \in \mathbb{R}^{C \times C}\), where each element \(T_{t, h} = P(y_i = h \mid \tilde{y}_i = t)\) represents the probability of an instance from true class \(t\) being mislabeled as class \(h\).  The probability of an instance from a tail class \(t\) being mislabeled as a head class \(h\) is relatively high,  such that \(T_{t, h} > T_{t, t'}\) for any other tail class \(t'\), where \(t'\) represents a rarer tail class with fewer instances than \(t\). In addition, the probability of a sample from a head class \(h\) being mislabeled as a tail class \(t\) is low, such that \(T_{h, t} \approx 0\). Since noise may change the order of class sizes, we let \(N_K\) denote the observed instance count of the class ranked \(k\)-th in size after sorting all classes in descending order by instance count. Consequently, we have \(N_1 \geq \tilde{N}_1\) and \(N_c \leq \tilde{N}_c\) as shown in Fig.~\ref{fig:description1}, while maintaining the overall decreasing order \(N_1 \geq N_2 \geq \cdots \geq N_c \). In this paper, our task is to learn a model \(\theta : x_i \rightarrow \tilde{y}_i\) that maps each validation instance \(x_i\) to its ground-truth label \(\tilde{y}_i\), using the tail-to-head long-tailed noisy label training dataset.

\section{Methodology}


 To address the issue, we propose Disentangling and Unlearning for Long-tailed and Label-noisy data (DULL) method. The key idea of DULL is to weaken and unlearn the salient incorrect features, thereby preventing the model from being misled. As illustrated in Fig.~\ref{method_all}, DULL comprises two main mechanisms: Inner-Feature Disentangling (IFD) and Inner-Feature Partial Unlearning (IFPU). IFD is designed to disentangle the highly entangled features internally. With the disentangled features, IFPU weakens the salient incorrect features regions, enabling the model to unlearn wrong knowledge and achieve robustness against noisy labels. The algorithm and the training process details can be found in Appendix.~\ref{sec:Training_details}.

\subsection{Inner-feature disentangling}
\label{Inter-Class Knowledge Decoupling}



Before unlearning, it is essential to disentangle features internally. The features channels are highly entangled across classes, meaning that a channel is associated with multiple classes. This entanglement poses a challenge when attempting to unlearn features, it risks impacting the feature regions associated with correct classes. 

To address this issue, we introduce the Inner-Feature Disentangling (IFD) mechanism. IFD is aimed to separate the features channels into independent, class-specific regions. This ensures that each channel is associated with only one class, allowing to selectively unlearn feature regions related to incorrect classes without impacting those related to correct classes.

IFD incorporates a channel-class correlation matrix \(G\). As shown in Fig.~\ref{method_all}, \(G \in \mathbb{R}^{K \times C}\) is a learnable parameter matrix, in which \(K\) is the number of channels of the final feature map, \(C\) is the number of classes. 
Each element \(G_{ij} \in [0,1]\) indicates the correlation between the \(i\)-th class and the \(j\)-th channel, where higher values represent stronger relevance.

We start by optimizing \(G\) to capture channel-class correlations \cite{lin2023erm}. For \((x_i, y_i)\), the \(y_i\)-th column of \(G\) serves as a mask multiplied onto the feature graph \(f(x_i)\) to shut down channels irrelevant to \(y_i\). This process outputs a masked feature graph \(\overline{f(x_i)}\). Then, both the \(f(x_i)\) and \(\overline{f(x_i)}\) are passed through the classifier. We employ the following loss to jointly optimize the standard classification and the masked classification of \(G\):
\begin{equation}
  L_0(f, \theta; G) = \sigma(y_i, \theta(f(x_i))+\sigma(y_i, \theta(\overline{f(x_i)}),
  \label{eq:equation_zero}
\end{equation}

where \(f\) denotes the backbone, \(\theta\) denotes the classifier and \(\sigma\) denotes the cross entropy loss. Then, an orthogonality regulation term is introduced to ensure the rows of \(G\) orthogonal, disentangling the channels:
\begin{equation}
  L_1(f, \theta; G) = \beta \| G^T G - I \|_F^2, I \in \mathbb{R}^{K \times K},
  \label{eq:equation_one}
\end{equation}

where \(\beta\) is a hyperparameter used to control the strength of regularization, \(I\) is an identity matrix, and \(\| \cdot \|_F\) denotes the Frobenius norm.
Finally, to promote the optimization of orthogonality, we incorporate a sparsity regulation term. 
The overall IFD optimization formula is as follows:
\begin{equation}
\begin{split}
  L_{IFD}(f, \theta; G)  = L_0 + L_1 + \|G\|_p,
  \label{eq:equation_IKD}
\end{split}
\end{equation}

where \(\| \cdot \|_p\) denotes the \(p\)-norm. 
Through this optimization, we obtain a \(G\) where each channel is linked to a single class, and each class is connected to at least one channel. This orthogonal structure guarantees that the deactivation of specific channels—essentially, the unlearning of certain feature regions—can not impact those related to correct classes.

\subsection{Inner-feature partial unlearning}
\label{Inner-Instance Partial Unlearning}



The impact of noisy labels on models lies in compelling the model to reinforce the incorrect feature representations connected to wrong classes, thereby causing the model to output incorrect predictions, as shown in Fig.~\ref{fig:motivation1}. 

To address this issue, we introduce the Inner-Feature Partial Unlearning (IFPU) mechanism. Based on the internally disentangled features from Sec.~\ref{Inter-Class Knowledge Decoupling}, IFPU identifies and unlearns feature regions associated with incorrect classes, thereby weakening the model's reinforcement of these features regions. 


First, IFPU identifies which regions of a single feature need to be deactivated and unlearned. We start by using a multi-label set $\mathcal{Y}_i$ gained in Sec.~\ref{Adaptive Fuzzy Multi-Labeling} for each instance $x_i$. The classes which are not included in $\mathcal{Y}_i$ represent the misleading or incorrect classes for $x_i$. With the classes to be unlearned identified, we compute an instance-level mask $M_i$ using \(G\) to determine which feature regions to be deactivated. $M_i$ is a binary vector of length $K$, where components with a value of 1 represent active channels (corresponding to correct classes), and components with a value of 0 mean inactive channels (corresponding to incorrect classes). The computation of $M_i$ is as follows:
\begin{equation}
  M_i(\mathcal{Y}_i, G) = \mathbb{I} \left( \sum_{j \in \mathcal{Y}_i} G_{j} > 0 \right),
  \label{eq:equation_mask}
\end{equation} 

where $\mathbb{I}(\cdot)$ is an indicator function and $G_{j}$ denotes the $j$-th column of $G$. 


Second, IFPU updates the model to achieve unlearning by minimizing the mean squared error (MSE) between the original prediction and the prediction from the masked features.
For \((x_i, y_i)\), the backbone $f$ extracts the feature map $f(x_i)$. We multiply the $M_i$ onto $f(x_i)$ to deactivate channels associated with incorrect classes, obtaining a masked feature graph $\overline{f(x_i)}$. 
Then, both the $\overline{f(x_i)}$ and the $f(x_i)$ are passed through the classifier $\theta$ and produce two predictions. We calculate the two predictions using MSE as follows:
\begin{equation}
  L_{IFPU}(f, \theta) = \text{MSE}(\theta(f(x_i), \theta(\overline{f(x_i)}).
  \label{eq:equation_IIPU}
\end{equation} 


Through IFPU, the model effectively unlearns and weakens the incorrect feature regions associated with wrong classes. This mechanism mitigates the misguidance caused by noisy labels and enhances the robustness of the model against such noise, improving classification performance.

\subsection{Adaptive multi-labeling}
\label{Adaptive Fuzzy Multi-Labeling}


Adaptive multi-labeling is crucial to both Sec.~\ref{Inner-Instance Partial Unlearning} and Sec.~\ref{Head-to-tail knowledge transfer}. It enables IFPU to identify misleading or incorrect classes. Additionally, it provides softened labels for the mixup operation during knowledge transfer, mitigating the negative impact of hard noisy labels.

The goal of adaptive multi-labeling is to output a labels set $\mathcal{Y}_i$ for each sample that includes the true label while excluding noisy ones. This is similar with the objective of noise detection, but different. The $\mathcal{Y}_i$ does not necessarily include all non-noisy labels. Instead, it focuses on providing a subset of labels that are most likely to represent the true class. In the T2H scenario, noisy labels mainly appear in head classes, which are relatively easier to identify and narrow the range of non-noisy labels. Here, we employ the Jensen-Shannon Divergence (JSD) \cite{karim2022unicon, lu2023label} to distinguish between noisy and clean samples and then construct $\mathcal{Y}_i$ for each instance. Other metrics, such as loss \cite{han2018co}, could also be used. How to separate noise is not our focus.


To obtain more diverse information from different perspectives, we first employ a dual-view strategy with weak and strong augmentations \cite{li2023disc, karim2022unicon, li2020dividemix}. The prediction for the weakly augmented view of \(x_i\) is denoted as \(p_w(x_{i})\), and the prediction for the strongly augmented view is \(p_s(x_{i})\). To leverage both views, we calculate fused prediction confidence \(p_{ws}(x_i)\) based on the \(p_w(x_{i})\) and \(p_s(x_{i})\):
\begin{equation}
\begin{split}
  p_{ws}(x_i) = \gamma \cdot p_w(x_i) + (1 - \gamma) \cdot p_s(x_i),
  \label{eq:equation_fuse}
\end{split}
\end{equation}

where \(\gamma\) is the fuse factor. 
We quantify the discrepancy $d_i \in (0,1)$ between the \(p_{ws}(x_i)\) and the label $y_i$ using the Jensen-Shannon Divergence (JSD) as follows:
\begin{align}
d_i = & \frac{1}{2} \text{KL}\left( y_i \,\middle\|\, \frac{y_i + p_{ws}(x_i)}{2} \right) \nonumber \\
& + \frac{1}{2} \text{KL}\left( p_{ws}(x_i) \,\middle\|\, \frac{y_i + p_{ws}(x_i)}{2} \right),
\end{align}

where $KL$ denoting the Kullback-Leibler divergence. A higher $d_i$ value indicates a greater discrepancy between the \(p_{ws}(x_i)\) and the $y_i$, while a lower $d_i$ value signifies better alignment. We treat \(d_i\) as a selected ratio. The count of the multi-label set \(q\) for \(x_i\) is calculated as follows:
\begin{equation}
\begin{split}
  q = \max(1, d_i \times C).
  \label{eq:equation_label_count}
\end{split}
\end{equation}

$\mathcal{Y}_i$ contains at least one label.
Following the \cite{karim2022unicon}, we compute the cutoff threshold and separate all samples into a clean set $\mathcal{D}_x$ and a noisy set $\mathcal{D}_u$. We then construct $\mathcal{Y}_i$ for each sample as follows:
\begin{footnotesize}
\begin{equation}
\mathcal{Y}_i = 
\begin{cases} 
\{y_k \mid p_{ws}(x_i, y_k) \in \text{Top-}q(p_{ws}(x_i))\}, & \text{if } x_i \in \mathcal{D}_x, \\
\{y_k \mid p_{ws}(x_i, y_k) \in \text{Top-}q(p_{ws}(x_i) \setminus \{y_i\})\}, & \text{if } x_i \in \mathcal{D}_u.
\end{cases}
\end{equation}
\end{footnotesize}
where $y_k$ is the $k$-th candidate label. For $x_i$ in $\mathcal{D}_x$, the $\mathcal{Y}_i$ is composed of the top $q$ labels with the highest confidence from the $p_{ws}$. For $x_i$ in $\mathcal{D}_u$, the $\mathcal{Y}_i$ is composed of the top $q$ labels with the highest confidence in the $p_{ws}$, excluding the noisy label $y_i$.

\subsection{Head-to-tail knowledge transfer}
\label{Head-to-tail knowledge transfer}

Through the above entire process, we have trained the model to avoid reinforcing incorrect features, thereby enhancing its robustness to noisy labels. However, the original distribution is still long-tailed. Therefore, we propose a knowledge transfer method that combines mixup and label smoothing to synthesize new samples, aiming to supplement the number of tail classes, further transferring knowledge from head to tail.

We first select samples pairs for mixing based on the similarity of the masked features $\overline{f(x_i)}$ in Sec.~\ref{Inner-Instance Partial Unlearning}. For each batch, we calculate the inner product between each samples pair to construct an inner product matrix. Next, we set the diagonal of the matrix to zero to avoid self-mixing and set the inner product with labels of higher-ranked classes to zero, preventing forward transfer to head classes. After normalizing the inner product matrix, we select samples pairs with high inner product values from different classes. 

Subsequently, to mitigate the negative impact of hard noisy labels, we use the $\mathcal{Y}_i$ obtained in Section~\ref{Adaptive Fuzzy Multi-Labeling} as a softener of the hard labels.  The normalized form of $\mathcal{Y}_i$ is denoted as $\hat{y}_i$, which is a vector of length $C$ with a sum equal to 1. We then combine these normalized labels with the original labels to generate a smoothed label for mixup:
\begin{equation}
  y'_{i} = (1 - \alpha) y_i + \alpha \hat{y}_i,
  \label{eq:equation_smooth}
\end{equation}
where \(\alpha\) is a smoothing factor that controls the degree of smoothing.The new instances are generated as follows:
\begin{equation}
  \tilde{x} = \lambda x_i + (1 - \lambda) x_j,
  \label{eq:equation_mix_x}
\end{equation}
\begin{equation}
  \tilde{y} = \lambda y'_{i} + (1 - \lambda) y'_{j},
  \label{eq:equation_mix_y}
\end{equation}
where \(\lambda \in [0, 1]\) is a mixing coefficient drawn from a Beta distribution, usually set as 0.5. 
The synthesized instances $(\tilde{x}, \tilde{y})$ supplement the number of tail classes and enhance the performance of the tail classes.

\section{Experiments}

\begin{table*}
    \centering
    \small
    \caption{Comparison of classification accuracy (\%) across long-tailed datasets with simulated T2H noise on CIFAR-10 and CIFAR-100. 
    The original imbalance factor (IF) is 10  (left of the arrows), with the new IF after the T2H noise introduction shown on the right, which exacerbates the imbalance. The best results are in \textbf{bold}.}
    \resizebox{1\linewidth}{!}{
    \setlength\tabcolsep{12pt}
    \begin{tabular}{l|l|cccc|cccc}
        \toprule
        \multirow{3}{*}{Dataset} & \multirow{2}{*}{Noise Ratio} & \multicolumn{4}{c|}{CIFAR-10} & \multicolumn{4}{c}{CIFAR-100} \\
        \cmidrule{3-10}
        & & T2H.10\% & T2H.20\% & T2H.30\% & T2H.40\% & T2H.10\% & T2H.20\% & T2H.30\% & T2H.40\% \\
        
        \cmidrule{2-10}
        & Imbalance Factor & $10 \rightarrow 12$ & $10 \rightarrow 15$ & $10 \rightarrow 20$ & $10 \rightarrow 25$ & $10 \rightarrow 15$ & $10 \rightarrow 20$ & $10 \rightarrow 30$ & $10 \rightarrow 40$ \\
        
        \midrule
        \multirow{1}{*}{Baseline} & CE & 75.45 & 72.42 & 65.82 & 59.43 & 48.31 & 46.57 & 41.51 & 34.64 \\
        \midrule
        \multirow{5}{*}{LT} 
        & LDAM \cite{cao2019learning} & 80.33 & 76.12 & 67.82 & 64.24 & 50.46 & 43.81 & 38.28 & 34.69 \\
        & LA \cite{menon2020long} & 65.37 & 65.73 & 60.52 & 54.14 & 36.14 & 31.17 & 27.37 & 21.41 \\
        & cmo \cite{park2022majority} & 77.68 & 78.81 & 71.60 & 66.06 & 50.72 & 47.06 & 41.78 & 38.59 \\
        & GCL \cite{li2022long} & 83.87 & 81.72 & 79.62 & 75.61 & 46.41 & 42.59 & 39.39 & 32.71 \\
        & DisA \cite{gao2024distribution} & 84.47 & 82.97 & 77.91 & 69.66 & 57.96 & 52.43 & 44.93 & 37.96 \\
        \midrule
        \multirow{5}{*}{NL} 
        & Co-teaching \cite{han2018co} & 82.64 & 55.34 & 37.65 & 29.61 & 46.99 & 36.95 & 26.93 & 17.39 \\
        & Co-learning \cite{tan2021co} & 85.28 & 82.75 & 77.26 & 67.08 & 56.79 & 51.69 & 46.75 & 40.25 \\
        & Mixup \cite{2017mixup} & 84.11 & 79.05 & 71.75 & 61.43 & 51.34 & 45.78 & 40.06 & 31.17 \\
        & GCE \cite{zhang2018generalized} & 85.51 & 79.25 & 71.52 & 62.98 & 48.21 & 42.41 & 33.17 & 30.67 \\
        & DivideMix \cite{li2020dividemix} & 73.74 & 73.91 & 75.41 & 74.39 & 49.87 & 48.51 & 46.79 & 40.55 \\
        & UNICON \cite{karim2022unicon} & 75.99 & 76.44 & 78.12 & 76.13 & 50.99 & 50.43 & 47.17 & 42.75 \\
        & JoCoR \cite{wei2020combating} & 70.28 & 51.84 & 36.73 & 24.62 & 46.35 & 39.01 & 28.63 & 17.82 \\
        \midrule
        \multirow{4}{*}{LTNL} 
        & HAR \cite{cao2020heteroskedastic} & 77.41 & 70.57 & 66.51 & 57.85 & 44.89 & 38.64 & 32.95 & 25.47 \\
        & RoLT \cite{wei2021robust} & 81.08 & 77.76 & 72.28 & 66.24 & 49.34 & 43.75 & 39.56 & 32.65 \\
        & RoLT-DRW \cite{wei2021robust} & 84.54 & 82.59 & 80.34 & 77.42 & 50.85 & 47.54 & 44.53 & 39.21 \\
        & TABASCO \cite{lu2023label} & 74.99 & 79.39 & 76.78 & 75.21 & 55.01 & 53.91 & 50.51 & 45.37 \\
        \midrule
        \multirow{1}{*}{Ours} & DULL & \textbf{86.49} & \textbf{84.25} & \textbf{81.53} & \textbf{80.43} & \textbf{59.98} & \textbf{55.12} & \textbf{52.43} & \textbf{46.48} \\
        \bottomrule
    \end{tabular}}
    \label{tab:simulated}
\end{table*}

\subsection{Datasets setup}

In order to comprehensively evaluate our method, we conduct experiments on both simulated and real-world long-tailed datasets with noisy labels respectively.

\noindent
\textbf{Simulated long-tailed datasets with T2H noise}. We investigate the existing noise addition types and find that none of them can simulate the T2H noise phenomenon. To provide a controlled experimental environment, we propose a noise addition algorithm that unidirectionally transfers tail class samples to head classes, simulating T2H noise. This dataset is constructed based on CIFAR-10 and CIFAR-100 with varying noise ratios and types. CIFAR-10 has 10 classes of images, including 50,000 training images and 10,000 testing images of size 32 × 32. CIFAR-100 has 100 classes, which contains 50,000 training images and 10,000 testing images. The details for the construction of simulated long-tailed datasets with T2H noise are as follows. 

\begin{table}[!t]
    \centering
    \caption{The classification accuracy (\%) on the test dataset of real-world T2H. The best results are in \textbf{bold}.}
    \resizebox{\linewidth}{!}{
    \begin{tabular}{lc|lc}
        \hline
        Method & Accuracy (\%) & Method & Accuracy (\%) \\ \hline
        CE & 26.88  & DisA \cite{gao2024distribution}  & 28.53  \\ 
        GCE \cite{zhang2018generalized}  & 21.05  & DivideMix \cite{li2020dividemix}  & 30.63  \\ 
        Mixup \cite{2017mixup}  & 28.93  & HAR \cite{cao2020heteroskedastic}  & 24.43  \\ 
        Co-teaching \cite{han2018co}  & 15.91  & RoLT \cite{wei2021robust}  &  22.67 \\ 
        Co-learning \cite{tan2021co}  & 31.51  & RoLT-DRW \cite{wei2021robust}  & 28.15  \\ 
        JoCor \cite{wei2020combating}  & 16.39  & TABASCO \cite{lu2023label}  & 31.44  \\ 
        cmo \cite{park2022majority}  & 29.71  & DULL(Ours)  & \textbf{33.61}  \\ \hline
    \end{tabular}
    }
    \label{tab:real-world_cifar100}
\end{table}

We start on a long-tailed dataset \(\tilde{\mathcal{D}}\) with an original imbalance factor (IF). The IF quantifies the degree of imbalance, defined as the ratio between the number of samples of the largest class and that of the smallest class. The class with the largest number of samples is identified as \(C_{max}\). We split \(\tilde{\mathcal{D}}\) into non-transferable set \(\mathcal{S}_0\) and transferable set \(\mathcal{S}\) (excluding \(C_{max}\)). The non-transferable set \(\mathcal{S}_0\) contains only samples of \(C_{max}\), and these samples cannot be transferred since there is no larger class for them to move to. \(\mathcal{S} = \{(x_i, \tilde{y}_i) | \tilde{y}_i \neq C_{max}\}\) contains all the samples that do not belong to \(C_{max}\). All noisy labels will be generated only in \(\mathcal{S}\). 
Next, we shuffle the transferable set \(\mathcal{S}\) and uniformly select a subset according to the noisy ratio \(r\), forming the preliminary noisy sample set \(\mathcal{S}' \in \mathcal{S}\). For each sample \((x_i, \tilde{y}_i) \in \mathcal{S}'\), we randomly generate a new noisy label \(y_i\) from the range \([0, \tilde{y}_i-1]\) as a sample from minor class can only be transferred to a larger class. The original label \(\tilde{y}_i\) is then replaced with the \(y_i\). This replacement is considered as a transfer of the sample from a minor class to a larger class.
Finally, we combine the processed \(\mathcal{S}\) with \(\mathcal{S}_0\) to get the T2H long-tailed noisy dataset \(\mathcal{D}\), completing the injection of noisy samples. The pseudocode for the construction method and the T2H noise addition algorithm are provided in the Appendix.~\ref{sec:dataset_construction}.

\begin{table}[!t]
    \centering
    \caption{The classification accuracy (\%) on the Clothing1M test dataset. The best results are in \textbf{bold}.}
    \resizebox{\linewidth}{!}{
    \begin{tabular}{lc|lc}
        \hline
        Method & Accuracy (\%) & Method & Accuracy (\%)  \\ \hline
        CE  & 68.94 & SL \cite{wang2019symmetric} & 71.02 \\
        Co-teaching \cite{han2018co} & 67.94 & GCE \cite{zhang2018generalized}  & 69.75 \\
        Dual-T \cite{yao2020dual} & 70.97  & Joint \cite{tanaka2018joint} & 72.23  \\
        PLM \cite{zhao2024estimating}  & 73.30 & DULL(Ours)  & \textbf{74.12}  \\ \hline
    \end{tabular}
    }
    \label{tab:clothing1m}
\end{table}

\noindent
\textbf{Real-world long-tailed datasets with noisy labels}. Real-world label-noisy datasets with long-tailed distribution adopted in our experiments include real-world T2H, Clothing1M~\cite{xiao2015learning} and WebVision-50. Real-world T2H is a long-tailed dataset with an original IF of 10 based on CIFAR-100, which is then re-labeled by model annotations to introduce noise. An example of the model annotation results is shown in \cref{fig:description1}, where the data distribution becomes more imbalanced compared to the original IF of 10. The evaluation is conducted on the CIFAR-100 test set. Clothing1M is a large-scale real-world benchmark widely used for label noise learning. It contains 1 million clothing images with noisy labels across 14 classes for training, with additional 14,000 samples for validation and 10,000 clean samples for testing. WebVision-50 is a long-tailed, noisy subset of the WebVision~\cite{li2017webvision} dataset, containing images from the first 50 classes for training, aligned with ImageNet ILSVRC12. Evaluation is conducted on both WebVision-50 validation set and the corresponding ILSVRC12 validation set. The noise rates in these real-world datasets are unknown, and inherent imbalances create challenging benchmarks for studying the combined impact of long-tailed distributions and label noise.

\subsection{Implementation details}

\noindent
\textbf{Compared methods}.We compare our method with the following three types of approaches: (1) Long-tail learning methods (LT)  include LDAM \cite{cao2019learning}, LA \cite{menon2020long}, CMO \cite{park2022majority}, GCL \cite{li2022long} and DisA \cite{gao2024distribution}; (2) Label-noise learning methods (NL) include Co-teaching \cite{han2018co}, Co-learning \cite{tan2021co}, Mixup \cite{2017mixup}, GCE \cite{zhang2018generalized}, DivideMix \cite{li2020dividemix}, UNICON \cite{karim2022unicon}, JoCoR \cite{wei2020combating}; (3) Methods designed for tackling long-tailed and label-noisy datasets (LTNL) include HAR \cite{cao2020heteroskedastic}, RoLT \cite{wei2021robust}, RoLT-DRW \cite{wei2021robust}, TABASCO \cite{lu2023label}.

\begin{table}[!t]
\centering
\caption{Ablation study on long-tailed CIFAR-10 and CIFAR-100 datasets with simulated T2H noise at 20\% and 40\% noise ratios. Results are shown under an original IF of 10.}
\resizebox{1.\linewidth}{!}{
    \begin{tabular}{ccc|cc|cc}
    \toprule
    \multirow{3}{*}{dual-views} & \multirow{3}{*}{IFPU} & \multirow{3}{*}{H2T.KT} & \multicolumn{2}{c|}{CIFAR-10} & \multicolumn{2}{c}{CIFAR-100} \\
    \cmidrule(lr){4-5} \cmidrule(lr){6-7}
     & & & \multicolumn{2}{c|}{ori.IF=10} & \multicolumn{2}{c}{ori.IF=10} \\
    \cmidrule(lr){4-5} \cmidrule(lr){6-7}
     & & & T2H. 20\% & T2H. 40\% & T2H. 20\% & T2H. 40\% \\
    \midrule
               &            &            & 72.42 & 59.43 & 46.57 & 34.64 \\
    \checkmark &            &            & 74.63 & 61.47 & 48.48 & 35.41 \\
               & \checkmark &            & 81.63 & 77.15 & 50.69 & 41.11      \\
    \checkmark & \checkmark &            & 83.24 & 79.43 & 51.66 & 42.35      \\
    \checkmark & \checkmark & \checkmark & \textbf{84.25} & \textbf{80.43} & \textbf{54.12} & \textbf{44.42} \\
    \bottomrule
    \end{tabular}
}
\label{tab:ablations}
\end{table}

\noindent
\textbf{Implementation details}. We employ ResNet-18 \cite{he2016deep} as the model for CIFAR-10 and CIFAR-100, while Pre-ResNet-34 \cite{he2016deep} is used for the Clothing1M and WebVision-50. The original models are trained for 200 epochs using Stochastic Gradient Descent (SGD) with an initial learning rate of 0.1, a momentum of 0.9, and a weight decay of $5 \mathrm{e}{-4}$. The learning rate is decayed by a factor of 10 at 100 and 150 epochs. We adopt a batch size of 1024. Moreover, the unlearned model is fine-tuned for 60 epochs using the same optimizer settings, with the learning rate decaying at epochs 10 and 20. Detailed hyperparameter settings ($\beta$, $\gamma$, $\alpha$, $\lambda$) and further experiments are provided in the Appendix.~\ref{sec:more_alations}.

\subsection{Experimental results}

\noindent
\textbf{Simulated long-tailed datasets with T2H noise}. Tab.~\ref{tab:simulated} reports the test accuracy of different types of methods on the long-tailed CIFAR-10/100 under the simulated T2H noise setting. The results reveal three main conclusions: (1) In most cases, existing long-tail methods (LT) fail to effectively handle T2H noise. This is mainly because the setting exacerbates the original imbalance ratio, causing these methods to overly focus on the tail while neglecting the head, and lack the ability to distinguish noisy samples effectively. (2) Label-noise methods (NL) and long-tail noisy labels (LTNL) methods show limited or completely fail in this setting. These methods face inherent limitations in detecting and correcting noisy labels, where direct corrections can introduce additional noise. (3) Our method outperforms other methods, demonstrating its effectiveness in mitigating inter-class entanglement and confusion caused by noisy samples and extracting core knowledge from the data. 

\noindent
\textbf{Real-world T2H}. The experimental results on the real-world long-tailed datasets with T2H Noise are detailed in Tab.~\ref{tab:real-world_cifar100}. Our approach consistently surpasses the existing baselines, achieving a notable 6.73\% improvement over the previous method. These results highlight the efficacy of our method in effectively managing noisy labels, particularly in scenarios involving challenging real-world T2H noise.

\noindent
\textbf{Clothing1M and WebVision-50}. The experimental results on the Clothing1M dataset are presented in Tab.~\ref{tab:clothing1m}. Compared to existing baseline methods, our proposed approach demonstrates improved performance on this dataset, achieving a 5.18\% improvement over the CE method. These results further validate the effectiveness and superiority of our method in handling complex real-world datasets. Results on the WebVision-50 dataset are provided in the Appendix.~\ref{sec:webvision50}

\subsection{Ablations and model validation}

\noindent
\textbf{Ablation studies on components of DULL}. As shown in Tab.~\ref{tab:ablations}, we conduct ablation studies on CIFAR-10/100 (IF = 10, T2H noise at 20\% and 40\%) to evaluate each module's impact. The core components tested are dual-view, IFPU, and H2T.KT.
The first row in the table presents the test accuracy of the baseline model. When the dual-view module was introduced, the test accuracy increased by 0.77\% to 2.21\%. This demonstrates that the dual-view captures diverse information, enhancing the model's generalization ability.
Adding the IFPU to the model further improved performance by 3.18\% to 17.96\%. This highlights the effectiveness of the IFPU, which efficiently filters out irrelevant and confusing knowledge, allowing the model to focus on the core feature information of the samples.
Incorporating the H2T.KT module led to an additional performance gain of 1\% to 2.46\%.

\begin{table}[!t]
    \centering
    \caption{OM and LSM values of DULL on long-tailed CIFAR-10 and CIFAR-100 datasets with simulated T2H noise at 20\%, 30\%, and 40\% noise ratios. The experiments are conducted with an initial imbalance factor (IF) of 10.}
    \resizebox{1.\linewidth}{!}{
    \begin{tabular}{l|ccc|ccc}
        \toprule
        \multirow{3}{*}{Dataset} & \multicolumn{3}{c|}{CIFAR-10} & \multicolumn{3}{c}{CIFAR-100} \\
        \cmidrule(lr){2-4} \cmidrule(lr){5-7}
        & \multicolumn{3}{c|}{ori.IF=10} & \multicolumn{3}{c}{ori.IF=10} \\
        \cmidrule(lr){2-4} \cmidrule(lr){5-7}
        & T2H.20\% & T2H.30\% & T2H.40\% & T2H.20\% & T2H.30\% & T2H.40\% \\
        \midrule
        OM & 1.17 & 0.93 & 1.39 & 28.91 & 20.26 & 27.49 \\
        LSM & 0.1055 & 0.1048 & 0.1076 & 0.0117 & 0.0113 & 0.0116 \\
        \midrule
    \end{tabular}
    }
    \label{tab:effective_ikd}
\end{table} 

\begin{table}[!t]
    \centering
    \begin{small}
    \caption{Classification accuracy across Head, Middle, Tail classes and Overall performance for different methods on CIFAR-100 with a simulated T2H noise ratio of 40\% and IF of 10. The best results are in \textbf{bold}.}
    \begin{tabular}{l|cccc}
        \hline
        Method & Head & Middle & Tail & Overall \\
        \hline
        CE & 48.71 & 41.31 & 28.22 & 35.92 \\
        RoLT \cite{wei2021robust} & 39.18 & 32.16 & 20.13 & 32.65 \\
        RoLT-DRW \cite{wei2021robust} & 38.75 & 35.25 & 24.23 & 39.21 \\
        DULL(Ours) & \textbf{62.10} & \textbf{50.13} & \textbf{32.49} & \textbf{46.48} \\
        \hline
    \end{tabular}
    \label{tab:fine-grained}
    \end{small}
\end{table}

\noindent
\textbf{Effectiveness of IFD}. To evaluate the effectiveness of the IFD in disentangling inter-class knowledge entanglement, we introduce two metrics for quantitative evaluation following \cite{lin2023erm}.
\begin{itemize}
    \item \textbf{Orthogonality measure (OM)}. OM quantifies the orthogonality between different classes by calculating the cosine similarity between the row of $G$.
\end{itemize}
\begin{itemize}
    \item \textbf{L1-Sparsity measure (LSM)}. LSM quantifies the sparsity of matrix $G$, capturing the degree of feature redundancy reduction.
\end{itemize}
A lower OM value indicates greater orthogonality between inter-class knowledge, effectively disentangling inter-class knowledge. A lower LSM value suggests reduced redundancy in $G$, allowing the model to focus on key channels relevant to each class. 
We evaluated OM and LSM for the IFD under different simulated T2H noise ratios using long-tailed CIFAR-10/100. As shown in Tab.~\ref{tab:effective_ikd}, the OM values of $G$ on CIFAR-10 and CIFAR-100 converge to low levels, indicating inter-class orthogonality and reduced entanglement. The LSM values convergence to $1/C$, indicating enhanced feature sparsity.

\noindent
\textbf{Effectiveness of multi-label in capturing true labels}. Our multi-label mechanism significantly improves the accuracy of corrected labels in matching true labels. Compared to semi-supervised methods, it better captures true labels and reduces reliance on incorrect labels. Experimental results, shown in Appendix.~\ref{sec:more_alations}

\noindent
\textbf{Performance across head, middle, and tail}. To evaluate effectiveness across different class types, we divided the dataset into Head, Middle, and Tail classes. As shown in Tab.~\ref{tab:fine-grained}, our method outperforms the baseline and other methods in all class types. 
Our approach achieves improvements in Tail classes while still enhancing accuracy for Head and Middle, resulting in better overall performance.

\section{Related work}
\label{sec:related work}

\subsection{Noisy labels learning on long-tailed data}

Detailed related work on long-tail learning and label-noise learning individually is provided in the Appendix.~\ref{sec:related_work_2}. Here, we focus on long-tailed noisy label learning, particularly addressing the challenge of identifying noisy labels in tail classes. Numerous studies have proposed specialized strategies to address this issue. RoLT introduces a prototype noise detection method based on class centroid distances \cite{wei2021robust}. ULC combines class-specific noise modeling while accounting for cognitive and incidental uncertainties \cite{huang2022uncertainty}. TABASCO employs a weighted JS divergence (WJSD) and adaptive centroid distance (ACD) to recognize clean samples from long-tailed noisy data \cite{lu2023label}.  These methods then commonly utilize semi-supervised learning to tackle identified noisy samples in the second correction stage. In addition, HAR proposes a heteroscedastic adaptive regularization method to handle noisy samples, applying higher intensity regularization to data points with high uncertainty and low density \cite{cao2020heteroskedastic}. RCAL utilizes representations extracted through unsupervised contrastive learning to eliminate noisy samples, restore the representation distribution, and further sample data points from this distribution to enhance the model's generalization ability \cite{zhang2023noisy}. 

\subsection{Machine unlearning}

Machine unlearning aims to selectively erase specific data points or classes from a model while preserving knowledge of the remaining data. Most methods adopt an approximate unlearning, where model parameters are fine-tuned to erase targeted information efficiently without requiring full retraining \cite{chen2023boundary, fan2023salun, neel2021descent, golatkar2020eternal, graves2021amnesiac, lin2023erm}. For example, SalUn~\cite{fan2023salun} calculates the saliency weights of the target forgetting dataset in the model and updates the model by removing these weights.
ERM-KTP introduces a knowledge-level unlearning framework that reduces class knowledge entanglement using a mask during training \cite{lin2023erm}. 
After receiving a forgetting request, the mask is used to transfer knowledge from non-target data points while prohibiting the knowledge of target points, enabling effective unlearning.

\section{Conclusion}

In this work, we have introduced DULL, a novel approach designed to tackle the challenges of long-tailed distributions and noisy labels, with a particular scenario on the "tail-to-head (T2H)" noise. Our method, comprising Inner-Feature Disentangling (IFD) and Inner-Instance Partial Unlearning (IFPU), mitigates the misguidance of noisy labels by unlearning incorrect feature regions. This process prevents the model from reinforcing wrong features, and enhances the model's robustness against noisy labels.
Extensive experiments on both simulated and real-world long-tailed datasets with noisy labels demonstrated the superior performance of our method compared to existing methods. However, the IFD may have limitations to optimize the matrix as the number of classes increases. Future work will focus on addressing these challenges to improve scalability and robustness.

{
    \small
    \bibliographystyle{ieeenat_fullname}
    \bibliography{main}
}

\clearpage
\setcounter{page}{1}
\maketitlesupplementary

\appendix
\section{Training details}
\label{sec:Training_details}

\begin{algorithm}[!t]
\caption{Training procedure of IKD}
\label{alg:ikd_alg}
\renewcommand{\algorithmicrequire}{\textbf{Input:}}
\renewcommand{\algorithmicensure}{\textbf{Output:}}
\begin{algorithmic}[1]
\REQUIRE Dataset $\mathcal{D} = \{(x_i, y_i)\}_{i=1}^N$, model backbone $\psi$, classifier $\theta$, learnable channel-class correlation matrix $G$.
\ENSURE Original model $(\psi, \theta, G)$.
\FOR{each epoch}
    \FOR{each batch in $\mathcal{D}$}
        \STATE Extract feature map $\psi(x_i)$.
        
        \STATE Compute masked feature map $\overline{\psi(x_i)}$.

        \STATE Compute original and masked predictions using Eq.~\ref{eq:equation_zero}.

        \STATE Compute orthogonality regularization $L_1$ using Eq.~\ref{eq:equation_one}.

        \STATE Compute sparsity regularization of $G$.
        
        \STATE Compute total loss $L_{IKD}$ using Eq.~\ref{eq:equation_IKD}.

        \STATE Update $G$, $\psi$, and $\theta$.
    \ENDFOR
\ENDFOR
\end{algorithmic}
\end{algorithm}

\begin{algorithm}[!t]
    \caption{Training procedure of IIPU}
    \label{alg:iipu_alg}
    \renewcommand{\algorithmicrequire}{\textbf{Input:}}
    \renewcommand{\algorithmicensure}{\textbf{Output:}}
    
    \begin{algorithmic}[1]
        \REQUIRE Dataset ${\mathcal{D}} = \{(x_i, y_i)\}_{i=1}^{N}$, 
        original model $(\psi, \theta, G)$, 
        unlearned model $(\Psi, \Theta)$.  
            
        \ENSURE Updated unlearned model $(\Psi, \Theta)$.  

        \FOR{each epoch}
            \FOR{each batch in \(\mathcal{D}\)}
                \STATE // Adaptive fuzzy multi-labeling
                
                \STATE   Generate dual-view predictions \(p_w(x_i)\), \(p_s(x_i)\).

                \STATE Calculate fused prediction confidence \(p_{ws}(x_i)\) using Eq.~\ref{eq:equation_fuse}.

                \STATE Compute instance fuzziness \(F(x_i)\) using Eq.~\ref{eq:equation_fuzziness}.
    
                \STATE Calculate adaptive multi-label count \(q\) using Eq.~\ref{eq:equation_label_count}.

                \STATE Assign fuzzy multi-label set \(\mathcal{Y}_i\).

                \vspace{1.0em}

                \STATE // Inner-instance partial unlearning

                \STATE Compute instance-level mask \(M(x_i)\) using Eq.~\ref{eq:equation_mask}.

                \STATE Extract feature maps \(\psi(x_i)\) and \(\Psi(x_i)\) from \(\psi\) and \(\Psi\).

                \STATE Apply \(M(x_i)\) to \(\psi(x_i)\) to shut down irrelevant channels.

                \STATE Partially unlearn within an instance using Eq.~\ref{eq:equation_IIPU}.

                \vspace{1.0em}

                \STATE // Head-to-tail knowledge transfer

                \STATE Calculate feature similarity within the batch.

                \STATE Select instance pairs with high similarity for mixing.

                \STATE Smooth multi-labels using Eq.~\ref{eq:equation_smooth}.

                \STATE Create new mixed instances using Eq.~\ref{eq:equation_mix_x} and Eq.~\ref{eq:equation_mix_y}.

            \ENDFOR
        \ENDFOR
        
        \RETURN Outputs
    \end{algorithmic}
\end{algorithm}

\begin{algorithm}[!t]
    \caption{T2H Dataset construction}
    \label{alg:construction_alg}
    \renewcommand{\algorithmicrequire}{\textbf{Input:}}
    \renewcommand{\algorithmicensure}{\textbf{Output:}}
    
    \begin{algorithmic}[1]
        \REQUIRE Long-tailed dataset $\tilde{\mathcal{D}} = \{(x_i, \tilde{{y}}_i)\}_{i=1}^{N}$, 
            noisy ratio $r$  
            
        \ENSURE Long-tailed dataset with T2H noise ${\mathcal{D}} = \{(x_i, y_i)\}_{i=1}^N$  

        \STATE  Find the category $C_{max}$ with the most samples in $\tilde{\mathcal{D}}$

        \STATE  Split $\tilde{D}$ into non-transferable set $S_0 = \{(x_i, \tilde{y}_i) \mid \tilde{y}_i = C_{\text{max}}\}$ and transferable set $S = \{(x_i, \tilde{y}_i) \mid \tilde{y}_i \neq C_{\text{max}}\}$.

        \STATE Shuffle $\mathcal{S}$ and uniformly select a subset $\mathcal{S}'$ according to noisy ratio  $r$.

        \FOR{$x_i, \tilde{y}_i \in \mathcal{S}'$}
            \STATE Randomly generate a noisy label $y_i$ in [0, $\tilde{y}_i-1$]
            \STATE Replace the original label $\tilde{y}_i$ with $y_i$
        \ENDFOR

        \STATE Combine $\mathcal{S}$ with $\mathcal{S}_0$ to form $\mathcal{D}$
        
        \RETURN Outputs
    \end{algorithmic}
\end{algorithm}

Our proposed method DULL is composed of two key steps, IKD and IIPU. The training procedures for the two steps are detailed in Alg.~\ref{alg:ikd_alg} and Alg.~\ref{alg:iipu_alg}, respectively. 
In Alg.~\ref{alg:ikd_alg}, IKD incorporates orthogonality and sparsity constraints to ensure effective knowledge disentanglement by optimizing a learnable channel-class correlation matrix.
Alg.~\ref{alg:iipu_alg} presents the process of IIPU, which selectively adapts and erases class-specific knowledge within an instance, enabling effective unlearning of noisy information.

\section{Simulated dataset construction}
\label{sec:dataset_construction}

To provide a controlled experimental platform, we propose a new noise addition algorithm to construct a simulated long-tailed dataset with T2H noise, as detailed in Alg.~\ref{alg:construction_alg}.

\section{WebVision-50 results}
\label{sec:webvision50}

The experimental results on the WebVision-50 dataset are presented in Tab.~\ref{tab:webvison}. Compared to existing methods, DULL demonstrates improvements, achieving a 12.09\% increase in accuracy on the WebVision-50 test dataset over ERM. Additionally, it achieves a 14.07\% improvement in accuracy on the ImageNet test dataset compared to ERM. These results show the robustness and effectiveness of DULL in addressing label noise and optimizing performance under challenging real-world conditions.

\section{More ablations and model validation}
\label{sec:more_alations}

To further validate the robustness and generalizability of our method, we conduct additional ablation studies across diverse settings, providing deeper insights into the contributions of key components.

\begin{table}[!t]
    \centering
    \caption{Test accuracy on WebVision-50 and ImageNet validation sets. The best results are in \textbf{bold}.}
    \label{tab:webvison}
    \small
    \begin{tabular}{lcc}
        \hline
        Train Method & WebVision-50 (\%) & ILSVRC12 (\%) \\
        \hline
        ERM & 62.50 & 58.50 \\
        Co-teaching & 63.58 & 61.48 \\
        INCV & 65.24 & 64.61 \\
        MentorNet & 63.00 & 57.80 \\
        CDR & 64.30 & 61.85 \\
        DULL & \textbf{74.89} & \textbf{72.57} \\
        \hline
    \end{tabular}
\end{table}

\begin{figure}[!t]
    \centering
    \begin{subfigure}[b]{0.46\linewidth}
        \includegraphics[width=\linewidth]{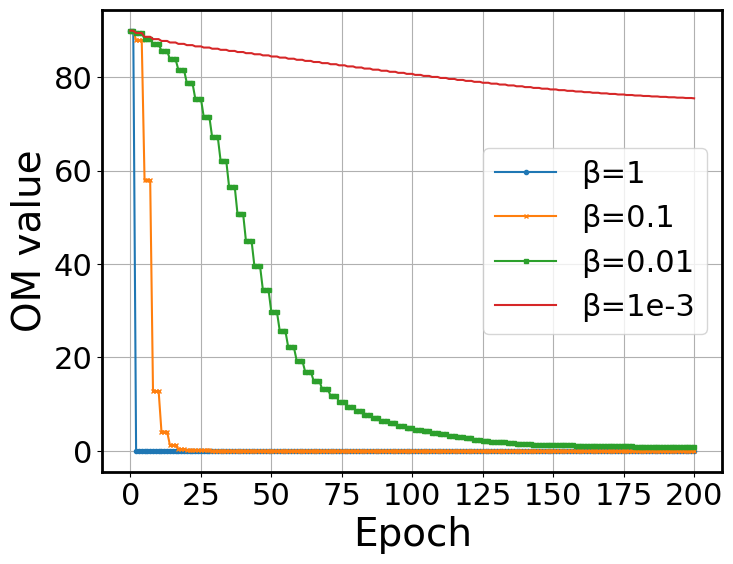} 
        \caption{OM value} 
        \label{fig:subfig1}
    \end{subfigure}
    \hfill
    \begin{subfigure}[b]{0.46\linewidth}
        \includegraphics[width=\linewidth]{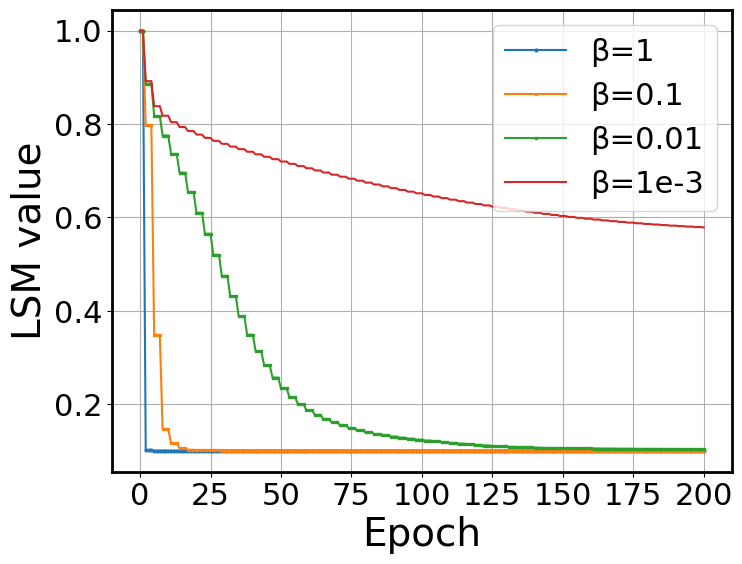} 
        \caption{LSM value} 
        \label{fig:subfig2}
    \end{subfigure}
    \caption{Hyperparameter sensitivity analysis of \(\beta\) in IKD on CIFAR10 with an original IF set to 10 and a simulated T2H noise ratio of 40\%.}
    \label{fig:hyper_beta}
\end{figure}

\begin{figure}[!t]
  \centering
   \includegraphics[width=0.5\linewidth]{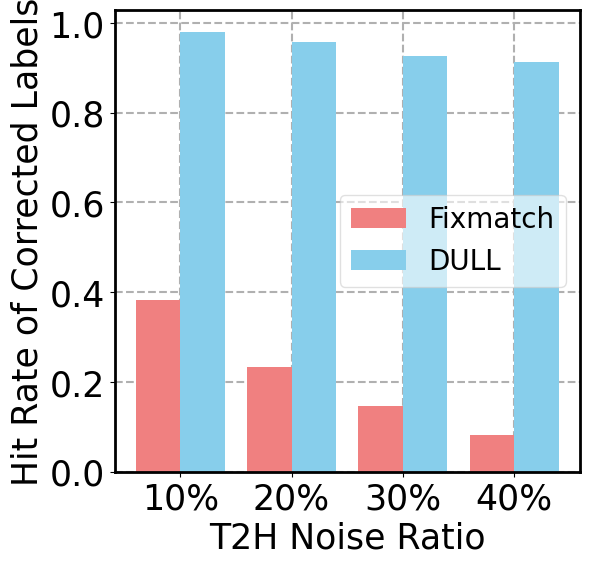}
   \caption{Hit rate of corrected labels for FixMatch and multi-label mechanism of DULL under different simulated T2H noise ratios on the long-tailed CIFAR-100 dataset with an original IF set to 10.}
   \label{fig:multi_label}
\end{figure}


\subsection{Sensitivity analysis on hyperparameter}

We explore the impact of hyperparameters (\(\gamma, \lambda, \alpha, \beta\)) through a detailed analysis. Here, we focus solely on the sensitivity analysis of \(\beta\),  while the other hyperparameters are set according to established conventions. The fuse factor \(\gamma\), commonly set to 0.5, represents an equal weighting between weak-augmented and strong-augmented predictions. The smoothing factor \(\alpha\) , typically set to 0.1, follows \cite{lukasik2020does}. The mixing coefficient \(\lambda\) is generally set to 0.5 for a balanced contribution from both instances. 

We conduct a sensitivity analysis on regularization strength \(\beta\) using the CIFAR10 with an original IF of 10 and a simulated T2H noise ratio of 40\%. This analysis aims to evaluate its influence on inter-class knowledge disentanglement and the enforcement of orthogonality. The values of \(\beta\) are configured as follows \{1, 0.1, 0.01, 0.001\}. As depicted in Fig.~\ref{fig:hyper_beta}, \(\beta = 1\) achieves the fastest convergence to near-zero levels for both OM and LSM, effectively enforcing orthogonality and sparsity . Conversely, smaller values of \(\beta\), such as \(\beta = 1\mathrm{e}{-3}\), result in slower convergence and a failure to converge. To balance convergence speed and stability, our method adopts \(\beta = 0.01\) as the default hyperparameter setting, ensuring robust performance in knowledge disentanglement.

\subsection{Effectiveness of multi-label in capturing true labels}

Fig.~\ref{fig:multi_label} illustrates the hit rate of corrected labels for FixMatch and the multi-label mechanism in DULL across varying simulated T2H noise ratios on the long-tailed CIFAR-100 dataset with an original IF of 10. As the noise ratio increases, FixMatch exhibits a consistent decline in performance, reflecting its limited ability to correct noisy labels under high noise levels. In contrast, the multi-label mechanism in DULL maintains a significantly higher hit rate, particularly under severe noise conditions. This demonstrates the ability of the multi-label mechanism in DULL to capture true labels, overcoming the limits of single-label correction.

\section{Related work part 2}
\label{sec:related_work_2}
\subsection{Long-tail learning}

Long-tail learning is a strategy aimed at improving the accuracy of tail classes while maintaining stable performance for head classes. Resampling is a classic method which adjusts the distribution of training data, primarily divided into oversampling \cite{nitesh2002smote} and undersampling. Reweighting adjusts the weights of samples during training to make the model pay more attention to tail classes. Depending on the weighting approach, it can be classified into loss function reweighting \cite{hong2021disentangling, ross2017focal, park2021influence, ren2020balanced, zhang2021distribution} and logit adjustment \cite{li2022long, menon2020long}. Recent studies have decoupled the model training process into two stages: the first stage focuses on training an effective feature extractor, while the second stage fine-tunes the classification \cite{kang2019decoupling, li2022long}. Additionally, data augmentation serves as a direct and effective method for generating and enriching tail class data by utilizing existing knowledge from head classes \cite{park2022majority, perrett2023use, wang2021rsg, zang2021fasa, li2021metasaug}. 

\subsection{Label-noise learning}

Label-Noise learning can be broadly categorized into two main directions, noisy label detection with correction, and robust label-noise learning. The detection and correction of noisy labels typically involve a two-step process: the first step identifies samples with incorrect labels using various metrics, and the second step corrects the labels of noisy samples. In the first stage, noise identification methods can be classified based on the metrics used, including loss-based methods \cite{han2018co, li2020dividemix} and JS divergence (JSD)-based methods \cite{karim2022unicon, lu2023label}. In the second stage, methods for handling noisy labels can include semi-supervised learning \cite{berthelot2019mixmatch, han2019deep, karim2022unicon, li2020dividemix, lu2023label}, sample re-weighting \cite{ren2018learning}, and label smoothing \cite{lukasik2020does}, among others. Traditional noisy label detection and correction methods have been effective in conventional settings. However, their performance declines in this study, particularly in accurately correcting noisy labels from tail classes, which can even introduce additional noise. In contrast, robust label-noise learning aims to mitigate or ignore the negative effects of noisy label samples by modifying the loss function, primarily through techniques such as regularization and loss correction \cite{cheng2022class, cheng2021mitigating, liu2020early, zhang2018generalized, lukasik2020does, xia2020robust}.

\end{document}